\documentclass[10pt,twocolumn,letterpaper]{article}
 
\usepackage{icb}
\usepackage{times}
\usepackage{epsfig}
\usepackage{graphicx}
\usepackage{booktabs}

\usepackage{amsmath}
\usepackage{amssymb}
 \usepackage{amsmath}
\usepackage{amsfonts}
\usepackage[colorlinks=true]{hyperref}
\usepackage{booktabs}

\usepackage{subfigure}
\usepackage{color,soul}
\renewcommand{\eqref}[1]{{{Eq.  \ref{#1}}}}
\newcommand{\figref}[1]{{{Figure \ref{#1}}}}
\newcommand{\secref}[1]{{{Section \ref{#1}}}}
\newcommand{\tabref}[1]{{{Table \ref{#1}}}}
\usepackage{multirow}
\usepackage[norule,symbol,perpage]{footmisc}

\ifdefined \submissioncopy
  \newcommand{\mh}[1]{}

\else
  \newcommand{\mh}{\textcolor{red}}

\fi

\usepackage{hyperref}
\hypersetup{
  pdfinfo={
    Title={Your Title Here},
    Author={Your Name Here},
    Subject={If you want to put something here, do so},
    Keywords={Add some keywords if you feel so inclined}
  }
}
\usepackage{pdfpages}

\icbfinalcopy 

\ificbfinal\fi

\makeatletter 
\def\ps@IEEEtitlepagestyle{ 
\def\@oddfoot{\mycopyrightnotice} 
\def\@evenfoot{} 
} 
\def\mycopyrightnotice{ 
{\hfill \footnotesize 978-1-7281-3640-0/19/\$31.00 \copyright 2019 IEEE\hfill} 
} 
\makeatother 

\begin{document}

\title{On the Effectiveness of Laser Speckle Contrast Imaging and Deep Neural Networks for Detecting Known and Unknown Fingerprint Presentation Attacks}

\author{Hengameh Mirzaalian$^{1}$, Mohamed Hussein$^{1,2}$, Wael Abd-Almageed$^{1}$\\
$^1$Information Sciences Institute, 4676 Admiralty Way \#1001, Marina Del Rey, CA 90292\\
$^2$Faculty of Engineering, Alexandria University, Alexandria, Egypt 21544\\
{\tt\small hengameh,mehussein,wamageed@isi.edu}
}

\maketitle
\thispagestyle{empty}


\begin{abstract}
Fingerprint presentation attack detection (FPAD) is becoming an increasingly challenging problem due to the continuous advancement of attack techniques, which generate ``realistic-looking" fake fingerprint presentations. Recently, laser speckle contrast imaging (LSCI) has been introduced as a new sensing modality for FPAD. LSCI  has the interesting characteristic of capturing the blood flow under the skin surface. Toward studying the importance and effectiveness of LSCI for FPAD, we conduct a comprehensive study using different patch-based deep neural network architectures. Our studied architectures include 2D and 3D convolutional  networks as well as a recurrent network using long short-term memory (LSTM) units. The study demonstrates that strong FPAD performance can be achieved using LSCI.  We evaluate the different models over a new large dataset. The dataset consists of 3743 bona fide samples, collected from 335 unique subjects, and 218 presentation attack samples, including six different types of attacks.  To examine the effect of changing the training and testing sets, we conduct a 3-fold cross validation evaluation. To examine the effect of the presence of an unseen attack, we apply a leave-one-attack out strategy. The FPAD classification results of the networks, which are separately optimized and tuned for the temporal and spatial patch-sizes, indicate that the best performance is achieved by LSTM.
\end{abstract}

\let\thefootnote\relax\footnotetext{\mycopyrightnotice} 
\section{Introduction}
\label{sec:Introduction}
Biometric authentication systems are widely used since they provide a higher level of security with a lower cost. They also eliminate the need to carry identifications cards or remember complicated passwords.  However, these systems are vulnerable to \textit{presentation attacks} (PA), i.e. presentation of fake biometric samples in order to impersonate an authorized user or obfuscate the identity of an illegal user. To  maintain the integrity of the biometric systems, developing accurate and robust  \textit{presentation attack detection} (PAD) techniques is essential.  However, due to the presence of realistic fake samples created by sophisticated attack-techniques, such as silicone and liquid latex, designing a successful PAD system is not trivial and PAD is becoming an increasingly challenging problem   \cite{Marcel2014:Handbook}. 

Fingerprint is perhaps the most conventional biometric identifier due to the long-time perception of its uniqueness, universality, measurability, and subject-friendliness \cite{JAIN2016:50Years}.  Fingerprint PAD (FPAD) methods make benefit of the  physiological measurement of the finger collected by the hardware-sensors \cite{Bowden-Peters2012:FoolingCapacitive}. These measurements are then used by a downstream algorithm (e.g. signal processing- or machine learning-based) to distinguish between bona fide (i.e. real) and PA samples.%
The  physiological measurements provided by the sensors can be either static or dynamic. Static characteristics, such as odor~\cite{Baldisserra2006:Odor}, skin resistance \cite{Kulkarni:2015:surveyFPDA}, and measurements by infrared  \cite{Steiner2016:MSI-SWIR-FacePADVer} and optical coherence tomography \cite{Cheng2006:OCT}, are extracted from a single fingerprint impression. On the other hand,  dynamic features are derived by processing multiple frames of the same fingerprint sample, e.g. a time series of  images  to measure elasticity \cite{Antonelli:2006:TIFS,Jia:ICB2007}, heartbeat \cite{Abhyankar:2009},   blood flow  \cite{vaz:16}, or   multi-spectral image acquisitions of finger under different illumination conditions using different wavelengths \cite{Maltoni:2009}. 
The aforementioned approaches are commonly referred to as \textit{hardware-based} techniques since they deploy additional hardware to the fingerprint sensing hardware. However, obviously, all such techniques involve both hardware and software components. Therefore, we may also refer to them as \textit{hybrid} techniques. This is in contrast to \textit{software-only} techniques, which are commonly referred to just as \textit{software-based} techniques. Software-only techniques do not augment fingerprint sensing hardware with any additional sensors and solely depend on the data used for recognition to perform fingerprint PAD \cite{Marasco2014:SurveyPADFinger,Sousedik2014:PADFingerSurvey}. The majority of the existing FPAD methods in this category apply traditional classification techniques  (e.g. support vector machine) using hand-crafted features (e.g. wavelet and  graylevel co-occurrence matrix of optical images) \cite{Derakhshani:2003:PR,Tan:2006:CVPR}. More recently, few FPAD approaches have been proposed  utilizing convolutional neural networks (CNNs). Nogueira et al. \cite{Nogueira:2016:IEEETIFS}  fine-tuned  AlexNet \cite{AlexNet:NIPS2012} and VGG  \cite{VGG:Simonyan14c} architectures to preform liveness detection of the fingerprints. A classical ConvNet consisting of four 2D convolutional layers with a binary cross-entropy loss was used by Wang et al. \cite{Wang:2015:ChineseConf} to do the FPAD task. Bhanu et al. \cite{Bhanu:2017:DL_bio_Book} used triplet loss in their network to minimize the intra-class distances of the patches belonging to the same class while maximizing the inter-class distances. Chugh et al. \cite{Chugh2017} used MobileNet-v1 over the  centered and aligned patches of the optical images to discriminate between fake and real fingerprints. Park et al. \cite{Eunsoo:2018} included  fire and gram modules within their network to learn the textures of the bona fide and PA samples. Kim et al. \cite{KIM:2016} employed deep belief networks and used contrastive divergence for FPAD. 
  
Software-only approaches are attractive due to their low cost and applicability on the widely used legacy fingerprint sensors. However, with the increased difficulty of FPAD, it is important to explore hybrid solutions that use extra dedicated sensors for FPAD. 
In this regard, laser speckle contrast imaging (LSCI) has attracted very little attention in the FPAD literature despite its interesting characteristics. LSCI  provides information on the flow of blood under the skin through the dynamics of speckle pattern. This information can be very valuable in discriminating between bona fide fingerprints and fake ones. 
Chatterjee et al. \cite{Chatterjee2017:antispoof} conducted a study over  the hardware and physics part of LSCI for FPAD and showed that there is a significant difference between the biospeckle patterns of the real fingers comparing to the patterns of the fake samples. However, the study was conducted on a very small sample and did not actually evaluate the performance of FPAD based on LSCI data.   Recently, Keilbach et al. \cite{Keilbach2018BIOSIG}  performed LSCI-based FPAD by applying an SVM classifier on a set of hand crafted features, such as intensity histograms and LBP features. Later on,  Hussein et al. \cite{wifs:Hussein:2018} applied a simplified version of AlexNet \cite{AlexNet:NIPS2012} to perform LSCI-based FPAD. Both studies showed promising performance of the proposed FPAD methods using LSCI data. However, the former only relied on hand crafted features while the latter deployed a network model that did not make the best use of the temporal characteristics of the LSCI data by using only 2D convolutions. Moreover, the evaluation dataset in both works was relatively small (only 163 subjects).

In this work, we conduct a comprehensive analysis of the LSCI-based FPAD utilizing deep neural networks.
Compared to the few earlier studies, our evaluation is conducted on a considerably larger dataset (335 subjects). Moreover, we explore different deep neural network architectures with the goal of making the best use of the spatial and temporal information in the LSCI data for FPAD. Our architectures include 2D and 3D spatial convolutional and recurrent models. We separately adapt the representation of the spatiotemporal LSCI data to the suitable format for each of the evaluated networks. All our models are patch-based, in which a classification score is estimated for each patch of the input sample, and the sample is classified based on averaging the patch-wise scores. We tune all the networks for the temporal and spatial patch-sizes.  Given the optimized models,  we first perform a 3-fold cross validation to examine the effect of changing the training and testing sets. Then, to examine the effect of the presence of an  unseen attack, which was not included within the training phase, we apply a leave-one-attack out evaluation strategy.  Cross validation results of the networks indicate that the best performance is achieved by the recurrent model and further inspection of the results indicates that the most challenging attack in our dataset is the dragon-skin overlay (\secref{sec:NumericalResults}).

The remainder of this paper is organized as follows. In \secref{sec:LSCI}, laser speckle contrast imaging and our capture device are briefly introduced. \secref{sec:Benchmark} discusses the proposed benchmark including details of the dataset, partitioning strategy and metrics used to evaluate various network architectures. In \secref{sec:Method}, we describe different deep network architectures investigated in this study. Evaluation results are presented in \secref{sec:NumericalResults}. Finally, conclusions are drawn in \secref{sec:Conclusions}.

\section{LSCI Capture Device}
\label{sec:LSCI}
When laser light illuminates a surface,  a random interference pattern will be formed by the reflected light, which is known as a speckle pattern.  The speckle pattern is affected by the roughness and/or  temperature of the surface.  In the presence of a stationary object, the pattern is static. However, the pattern will be changed  over time when there exists motion on the illuminated object, such as the movements of the blood cells under the skin surface. In fact, using laser speckle contrast imaging (LSCI), the blood perfusion of the tissue can be visualized  \cite{LSCI2013Briers}. By collecting a time-series of the LSCI data, the flow of the blood cells is detectable. Therefore, the LSCI measurements of the fingers constitute a useful liveliness signal for FPAD.
 
\begin{figure}
    \centering
    \subfigure[Schematic of our LSCI capture device]{\label{fig:device}\includegraphics[width=7cm]{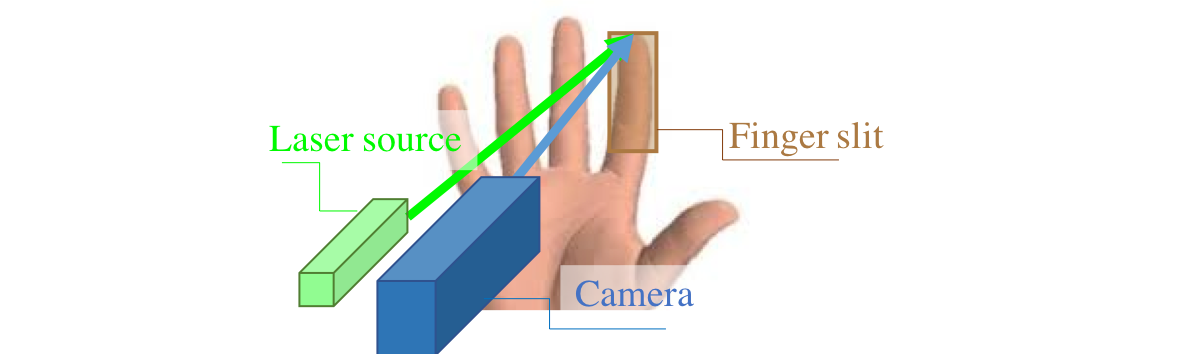}}\\ \hfill
    \subfigure[Capture area and LSCI data]{\label{fig:LSCI_ROI}\includegraphics[width=7cm]{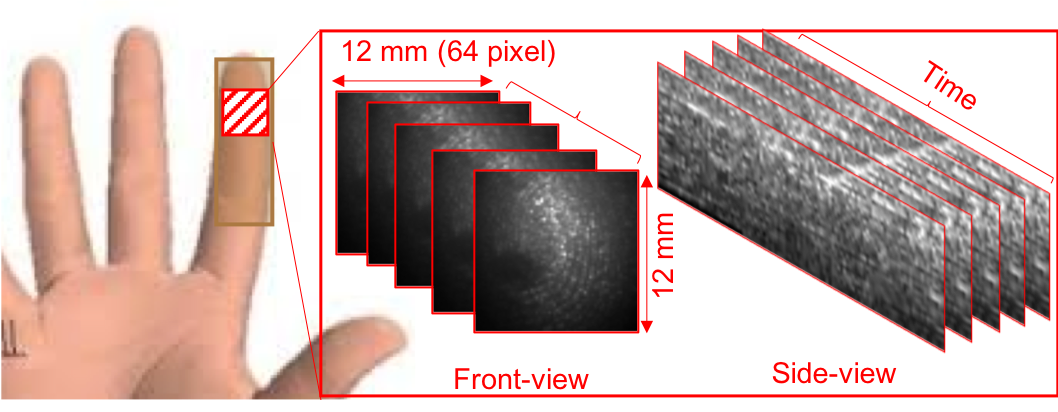}}\hfill
    \caption{(a) A simplified illustration of the LSCI  capture device, which consists of a laser source directed to the finger slit and a camera to capture the illumination reflection. (b) Area of the finger captured by the device is shown in hatched red.
    The middle and right images show the stacks of the 2D slices of the front-view and side-view of the LSCI data when viewed as a volume.
    }
\end{figure}

Our LSCI capture device consists of a 1310 nm laser illumination source and a camera. The camera constitutes an InGaAs area sensor, which is sensitive to the 1310 nm wavelength and not sensitive to visible light. The camera is also equipped with a lens assembly. The laser beam is produced by the laser source and directed  to a finger slit of dimensions $45$ mm $\times$ $15$ mm.  
A schematic diagram of the device is shown in \figref{fig:device}. No platen is used to cover the finger slit in our LSCI capture device. This shows that LSCI can be used in touch-less fingerprint sensors, which are generally less vulnerable to fingerprint lifting and are more desirable from the hygiene perspective \cite{Parziale2008:Touchless}. 

For this study, the optical design is adjusted to capture a region of size $12$ mm $\times$ $12$ mm that lies roughly in the middle of the first knuckle when the finger is placed on the slit while its tip is resting on the slit's edge, as shown in \figref{fig:LSCI_ROI}.  This region was found to always contain either skin for bona fide fingers or other material for PAs. The deployed InGaAs sensor is a low resolution ($64\times 64$ pixels) and high frame rate (up to $1000$ fps). In this study, the sensor is operated at a frame rate of approximately $500$ fps.

\section{Evaluation Benchmark}
\label{sec:Benchmark}
\subsection{Dataset}
\label{sec:Dataset}
Using the LSCI capture device explained in \secref{sec:LSCI}, we collected a large dataset of LSCI data for fingerprints. For each captured fingerprint, a sequence of $1000$ frames with the laser illumination are collected and $20$ frames without the laser illumination. The average of the frames captures without illumination is used for dead pixel and ambient illumination removal.

The LSCI data was collected from six fingers (the three middle fingers of each hand) of 335 unique subjects. Each subject was allowed to participate in the collection multiple times (up to three). At each participation, the subject passed by the collection station either in the absence of any attack or in the presence of up to two overlaid attacks attached to two of his/her fingers. The data was thoroughly reviewed by the research team. Samples with defects, \eg due to finger motion or hardware failure, were excluded. After this process, the dataset consisted of 3743 valid bona fide and 218 valid PA samples. The PA categories and number of collected samples per class are provided in \tabref{tab:InfDataset}. All used attacks are of overlay type. Two attack species have conductive coating material, one on a paper print and one on silicone. The other four attack species include two types of silicone, transparency print, and dragon-skin.

\begin{figure}[thb]
\centering 
\subfigure[Bona fide]{
\includegraphics[width=3.6cm]{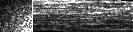}
\includegraphics[width=3.6cm]{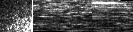}
}\\ [-0.7ex]
\subfigure[Conductive paper    ]{
\includegraphics[width=3.6cm]{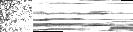}
\includegraphics[width=3.6cm]{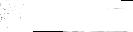}
}\\ [-0.7ex]
\subfigure[Conductive silicone]{
\includegraphics[width=3.6cm]{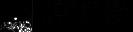}
\includegraphics[width=3.6cm]{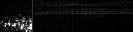}
}\\ [-0.7ex]
\subfigure[Transparency]{
\includegraphics[width=3.6cm]{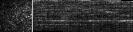}
\includegraphics[width=3.6cm]{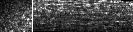}
}\\ [-0.7ex]
\subfigure[Silicone I]{
\includegraphics[width=3.6cm]{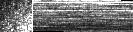}
\includegraphics[width=3.6cm]{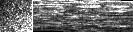}
}\\ [-0.7ex]
\subfigure[Silicone II]{
\includegraphics[width=3.6cm]{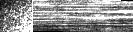}
\includegraphics[width=3.6cm]{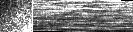}
}\\ [-0.7ex]
\subfigure[Dragon-skin]{
\includegraphics[width=3.6cm]{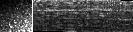}
\includegraphics[width=3.6cm]{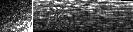}
}
\caption{ Sample LSCI data in the absence (a) and presence of PAs (b-g).  Each subfigure represents the front-view, i.e. first frame (64$\times$64), and side-view, i.e. frames border (64$\times$100), of a 100-frame LSCI data when viewed as a volume, for two different subjects within each category (a-g).  All pixel values in each volume are normalized to cover the display's dynamic range for visualization purposes.}
\label{fig:sampleLSCI}
\end{figure}

Examples of the acquired LSCI data in the presence and absence of the PAs are shown in  \figref{fig:sampleLSCI}. As can be seen in the side-view visualization, the changes in the pixel intensity over time is more noisy in the case of bona fide samples, which is due to the more dynamic speckle pattern, which is in turn due to the blood flow.

\subsection{Data Partitioning}
\label{sec:DataPartitioning}
We employ two different data partitioning strategies. In the first strategy, we use a 3-Fold partitioning to alleviate the bias resulting from a fixed division of the dataset into training, testing, and validation sets. The dataset is divided into three roughly equal sets of samples such that the data of each subject only appears in one set. Then, we create the 3-Fold partitioning by using two sets for training and one set for testing each time. $20\%$ of the training data is separated to create a validation set such that the data of each subject either appear all in the training or the validation set in each fold. As shown in \tabref{tab:3folds}, the distribution of the  bona fide and different PA samples  are approximately the same among the three folds.

\begin{table}
\centering
\footnotesize
\begin{tabular}{  l l l l  }  
Attack-type & \#  & Attack-type & \#  \\ \cmidrule(lr){1-2}\cmidrule(lr){3-4} 
 Conductive paper &  11  & Conductive silicone & 62  \\    
 Transparency   & 26  &  Silicone-I   &  13  \\   
 Silicone-II             & 79 &   Dragon-skin  & 27 \\    
\end{tabular}
\caption{ The number of the collected images at each of the PA categories in our dataset.} 
\label{tab:InfDataset}
\end{table}  
  
In order to evaluate the ability to detect previously \emph{unknown attacks} (i.e. attacks that were present in the training data),  we use a leave-one-attack-out (LOAO) data partitioning strategy. Since the dataset consists of six PA species (\secref{sec:Dataset}), we divide the dataset into six folds this time. The bona fide samples in the training, testing, and validation sets are fixed in all the folds. The training and validation sets of each fold include PA samples of only five of the PA species while all the samples the remaining PA species are put in the testing set. For example, the training set of Fold\#0 does not have any sample belonging to the conductive paper species, where the testing set of that fold  contains all the conductive paper samples. \tabref{tab:LOO} shows the distribution of the data samples per fold of the LOAO partitioning.


\subsection{ Evaluation Metrics}
\label{sec:Metrics}
Let  $P$ and $N$ represent the total numbers of the attack (positive) and bona fide (negative) testing sample, respectively. Given the number of false positives, $FP$, and number of false negatives,  $FN$, in the FPAD classification results of a given trained model, we measure the performance of the model in terms of the following metrics:  {i) bona fide presentation classification error rate $BPCER=FP/N$; ii) attack presentation classification error rate $APCER=FN/P$;  iii) average classification error rate $ACER=0.5(BPCER+APCER)$; iv) $BPCER$ at $APCER$ of $5\%$ denoted by $BPCER20$; and v) area under the Receiver Operating Characteristic (ROC) curve, $AUC$.


\begin{table*}[thp]
\centering
\footnotesize
\begin{tabular}{r c c c c c c c c c c c c c c c c c cc c c} \toprule
Folds&\multicolumn{3}{c}{Bona fide} &  \multicolumn{3}{c}{Conductive paper } & \multicolumn{3}{c}{Conductive silicone} & \multicolumn{3}{c}{Transparency} & \multicolumn{3}{c}{Silicone-I} & \multicolumn{3}{c}{Silicone-II } & \multicolumn{3}{c}{Dragon-skin} \\ \cmidrule(lr){1-1}\cmidrule(lr){2-4}\cmidrule(lr){5-7}\cmidrule(lr){8-10}\cmidrule(lr){11-13}\cmidrule(lr){14-16}\cmidrule(lr){17-19}\cmidrule(lr){20-22}
&\tiny{train} & \tiny{test}& \tiny{val} & 
\tiny{train} & \tiny{test}& \tiny{val} & 
\tiny{train} & \tiny{test}& \tiny{val} & 
\tiny{train} & \tiny{test}& \tiny{val} & 
\tiny{train} & \tiny{test}& \tiny{val} & 
\tiny{train} & \tiny{test}& \tiny{val} &
\tiny{train} & \tiny{test}& \tiny{val} \\
Fold\#0 &  1986  &  1240  &  517  &  6  &  3  &  2  &  34  &  20  &  8  &  14  &  9  &  3  &  42  &  26  &  11  &  7  &  4  &  2  &  14  &  9  &  4  \\
Fold\#1 &  1985  &  1254  &  504  &  6  &  4  &  1  &  33  &  21  &  8  &  14  &  8  &  4  &  42  &  27  &  10  &  7  &  4  &  2  &  14  &  9  &  4  \\
Fold\#2 &  1978  &  1249  &  516  &  6  &  4  &  1  &  33  &  21  &  8  &  14  &  9  &  3  &  42  &  26  &  11  &  6  &  5  &  2  &  14  &  9  &  4  \\ \toprule
\end{tabular}
\caption{Distribution of the data per fold of the  3-Fold partitioning strategy (\secref{sec:DataPartitioning}). }  
\label{tab:3folds}
\end{table*}

\begin{table*}[thp]
\centering
\footnotesize
\begin{tabular}{r c c c c c c c c c c c c c c c c c c c c c} \toprule
Folds&\multicolumn{3}{c}{Bona fide} &  \multicolumn{3}{c}{Conductive paper } & \multicolumn{3}{c}{Conductive silicone} & \multicolumn{3}{c}{Transparency} & \multicolumn{3}{c}{Silicone-I} & \multicolumn{3}{c}{Silicone-II } & \multicolumn{3}{c}{Dragon-skin} \\ \cmidrule(lr){1-1}\cmidrule(lr){2-4}\cmidrule(lr){5-7}\cmidrule(lr){8-10}\cmidrule(lr){11-13}\cmidrule(lr){14-16}\cmidrule(lr){17-19}\cmidrule(lr){20-22}
&\tiny{train} & \tiny{test}& \tiny{val} & 
\tiny{train} & \tiny{test}& \tiny{val} & 
\tiny{train} & \tiny{test}& \tiny{val} & 
\tiny{train} & \tiny{test}& \tiny{val} & 
\tiny{train} & \tiny{test}& \tiny{val} & 
\tiny{train} & \tiny{test}& \tiny{val} &
\tiny{train} & \tiny{test}& \tiny{val} \\
 Fold\#0 &  \color{black}3476  &  \color{black}79  &  \color{black}188  &  \color{red}0  &  \color{black}11  &  \color{black}0  &  \color{black}58  &  \color{black}0  &  \color{black}4  &  \color{black}24  &  \color{black}0  &  \color{black}2  &  \color{black}75  &  \color{black}0  &  \color{black}4  &  \color{black}12  &  \color{black}0  &  \color{black}1  &  \color{black}25  &  \color{black}0  &  \color{black}2  \\
Fold\#1 &  \color{black}3476  &  \color{black}79  &  \color{black}188  &  \color{black}10  &  \color{black}0  &  \color{black}1  &  \color{red}0  &  \color{black}62  &  \color{black}0  &  \color{black}24  &  \color{black}0  &  \color{black}2  &  \color{black}75  &  \color{black}0  &  \color{black}4  &  \color{black}12  &  \color{black}0  &  \color{black}1  &  \color{black}25  &  \color{black}0  &  \color{black}2  \\
Fold\#2 &  \color{black}3476  &  \color{black}79  &  \color{black}188  &  \color{black}10  &  \color{black}0  &  \color{black}1  &  \color{black}58  &  \color{black}0  &  \color{black}4  &  \color{red}0  &  \color{black}26  &  \color{black}0  &  \color{black}75  &  \color{black}0  &  \color{black}4  &  \color{black}12  &  \color{black}0  &  \color{black}1  &  \color{black}25  &  \color{black}0  &  \color{black}2  \\
Fold\#3 &  \color{black}3476  &  \color{black}79  &  \color{black}188  &  \color{black}10  &  \color{black}0  &  \color{black}1  &  \color{black}58  &  \color{black}0  &  \color{black}4  &  \color{black}24  &  \color{black}0  &  \color{black}2  &  \color{red}0  &  \color{black}79  &  \color{black}0  &  \color{black}12  &  \color{black}0  &  \color{black}1  &  \color{black}25  &  \color{black}0  &  \color{black}2  \\
Fold\#4 &  \color{black}3476  &  \color{black}79  &  \color{black}188  &  \color{black}10  &  \color{black}0  &  \color{black}1  &  \color{black}58  &  \color{black}0  &  \color{black}4  &  \color{black}24  &  \color{black}0  &  \color{black}2  &  \color{black}75  &  \color{black}0  &  \color{black}4  &  \color{red}0  &  \color{black}13  &  \color{black}0  &  \color{black}25  &  \color{black}0  &  \color{black}2  \\
Fold\#5 &  \color{black}3476  &  \color{black}79  &  \color{black}188  &  \color{black}10  &  \color{black}0  &  \color{black}1  &  \color{black}58  &  \color{black}0  &  \color{black}4  &  \color{black}24  &  \color{black}0  &  \color{black}2  &  \color{black}75  &  \color{black}0  &  \color{black}4  &  \color{black}12  &  \color{black}0  &  \color{black}1  &  \color{red}0  &  \color{black}27  &  \color{black}0  \\ \toprule
\end{tabular}
\caption{Distribution of the data per fold of the  LOAO partitioning strategy (\secref{sec:DataPartitioning}). The red-color encoded zeros indicate  the type of the PA excluded from the training-set at each fold (each row), e.g. Fold\#0 (first row) does not contain any image in the presence of conductive-print PA within the training data.}  
\label{tab:LOO}
\end{table*}

\section{Patch-based FPAD Techniques for LSCI}
\label{sec:Method}
To perform FPAD  using LSCI data, we investigate several patch-based neural network architectures. Based on the way deep networks consume LSCI data, these architectures can be categorized into three main groups: i) 2D architecture; ii) 3D architecture; and iii) temporal architecture. Since we are essentially using a new sensing modality (i.e. LSCI) and obtaining large amounts of training data is not feasible, we adopt a patch-based approach, in which we feed the neural network the LSCI patches, instead of the entire LSCI image. The architectural details  of each of the studied architectures and the patch sampling technique used for training them are explained in the following sections.

\subsection{Deep Neural Network Architectures}
\label{sec:NetModels}
We use the architecture proposed by Hussein et al. \cite{wifs:Hussein:2018} as our baseline network (BaseN). As shown in \figref{fig:BaseN}, BaseN consists of six consecutive 2D convolution (Conv) layers, where each 2D Conv is connected to a ReLu module. The output of the last 2D Conv layer is passed to a fully connected layer followed by a sigmoid layer to predict the PA probability score for the given input image patch.  
 
As depicted in \figref{fig:ResN}, the second network (ResN) is constructed by adding residual connections between every two 2D convolution layers of the BaseN. To make sure that the number of channels in the residual branch match the output of the main branch in each residual block, a 2D convolution layer is added to each residual connection. 

Our  third network (\figref{fig:IncpN}) contains inception modules (IncpN). The structure of this network is inspired by the GoogLeNet \cite{GoogleNet}. However, to account for the relatively limited number of LSCI data samples, we make our IncpN  model as a shallower version of GoogLeNet. 

To capture discriminative features along both spatial and temporal dimensions of the LSCI data, the forth network (Conv3) is made as a 3D version of the BaseN model by using 3D convolution and pooling layers instead of the 2D ones used in BaseN. As shown in \figref{fig:Conv3}, the 3D filters are of size $5\times 3 \times 3$ and the number of the filters at each convolution layer is set similar to those of BaseN.

The last network is made of a double-layer long short-term memory (LSTM) units with a hidden state of size 100. Compared to the other networks, the input data to the LSTM units has to be 1D vectors. Therefore, the input images are vectorized in this case, as shown in \figref{fig:LSTM_pipe} and explained in \secref{sec:Patch-based}.

The predicted PA probability score provided by each of the applied CNN networks  (\figref{fig:networks}) falls in the range $[0,1]$, where $1$ indicates that the input patch belongs to a PA region. To make a prediction for an entire input sample, the average of patch scores is used and a threshold  of $0.5$ is used to make the final predictions, i.e. the presence or absence of any PA within the image. 

All of the studies architectures  are optimized using Adam with binary cross entropy  loss and a learning rate of $2\times10^{-4}$. They are trained within 50 epochs with a batch size of 64. The final model is set  to be the one that minimizes the loss in the validation data.

\subsection{Patch Generation}
\label{sec:Patch-based}
Having access to a large enough number of training samples is a critical issue for training deep neural networks. Given that our LSCI dataset is  limited to 3957 images, similar to the prior FPAD works \cite{wifs:Hussein:2018,Bhanu:2017:DL_bio_Book,Nogueira:2016:IEEETIFS}, we adopt a patch-based approach in our pipeline. 

Each input sample is split into a set of small patches by sliding a window over the region of interest (ROI). In our implementation, the region of interest was set to be the central 32$\times$32 pixel region in each LSCI frame, which was found to enjoy the maximum intensity of the laser illumination in our experimental setup.  The deep networks are then trained over the generated patches. Given a test sample, patches are generated in the same way, and the classification scores are computed for each. The final score of the entire test sample is set as the average score over all patches.

The generated patches are 3D tensors of size $h\times w\times t$, where $h \times w$ are the spatial patch dimensions and  $t$ is the temporal dimension of the LSCI data. The temporal dimension of the patch is interpreted as the input channels by the 2D-model networks (BaseN, ResN, IncpN in \secref{sec:NetModels}), whereas it is counted as the third dimension of a 3D volume by the 3D-model network (Conv3  in \secref{sec:NetModels}). Suppose the patch's tensor is represented as a collection of 2D slices over time, $[\mathbb{F}_{h\times w}^1\mathbb{F}_{h\times w}^2...\mathbb{F}_{h\times w}^t]$. Feeding a patch to a 2D- or a 3D-model network is straightforward. However, for the LSTM network, a mapping function, $\Pi: \mathbb{F}_{h\times w}^i \mapsto \vec{v}^i$, has to be applied on each 2D slice of the temporal data $\mathbb{F}_{h\times w}^i$ to map it to a 1D vector. Typically, a convolutional network can be used to perform this mapping, e.g. \cite{DonahueCVPR2015}. However, in our case, due to the small input size, we adopt simple reshaping in our system.
Therefore, each  2D frame $\mathbb{F}_{h\times w}^i$   is mapped to a vector of size $hw \times$1 denoted by $\mathbb{F}_{hw\times 1}^i$. 

As shown in \figref{fig:LSTM_pipe}, the input to the LSTM network is $t$ vectors of size $hw \times$1: $[\mathbb{F}_{hw\times 1}^1\mathbb{F}_{hw\times 1}^2 ... \mathbb{F}_{hw\times 1}^t]$.    In our implementation,  the output of the last hidden state is passed to the fully connected layer at the end of the network.

In \secref{sec:NumericalResults}, we empirically evaluate the effect of the spatial and temporal patch-sizes  by testing the networks over: $h\times w \in [8\times 8,16 \times 16,32 \times 32,64 \times 64]$   spatial-patch sizes and $t \in [5,10,50,100]$ temporal-patch-sizes.

\begin{figure*}[htb]
\centering
   \subfigure[\vspace{-1.1ex }BaseN]{\label{fig:BaseN}\includegraphics[width=16.2cm]{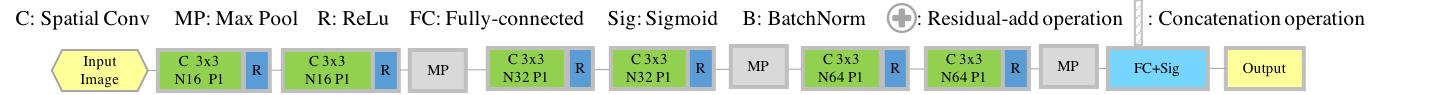}}\\ [-1.1ex]
   \subfigure[ResN]{\label{fig:ResN}\includegraphics[width=15.0cm]{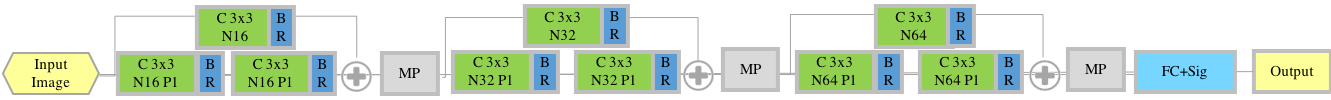}}\\ [-1.3ex]
    \subfigure[IncpN]{\label{fig:IncpN}\includegraphics[width=16.2cm]{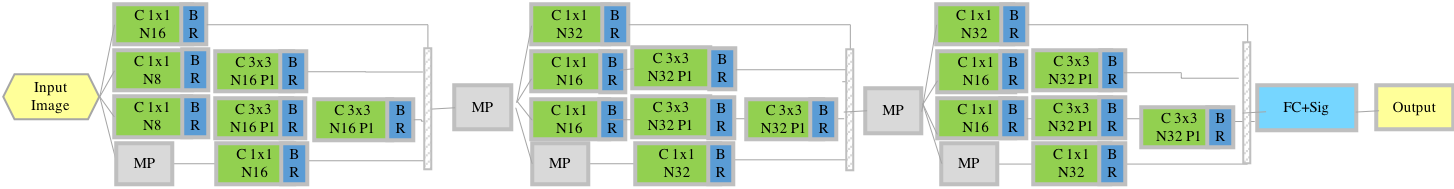}}\\ [-1.3ex]
    \subfigure[Conv3]{\label{fig:Conv3}\includegraphics[width=15.2cm]{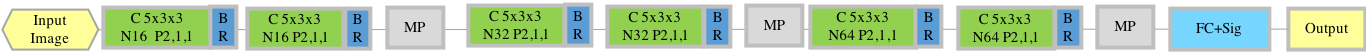}}\\ [-1ex]
    \subfigure[LSTM]{\label{fig:LSTM}\includegraphics[width=9.6cm]{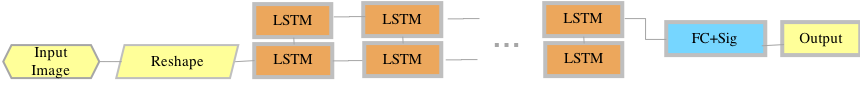}}
    \caption{   Different network structures applied to perform FPAD over LSCI data (\secref{sec:NetModels}). Depending on the way these models are incorporating the temporal dimension of the LSCI data to the network, they can be categorized to: i) 2D-model (BaseN, ResN, IncpN); ii) 3D-model (Conv3); and iii) Temporal-model (LSTM). The parameters of the convolutional layer are provided in front of the C, N, and P characters, e.g. C 3x3: kernel-size of 3x3, N16:  out-channel size of 16; and P1: padding of size 1 and the default padding-size is 0.}
    \label{fig:networks}
\end{figure*}

\begin{figure}[t]
    \centering
    \includegraphics[width=7cm]{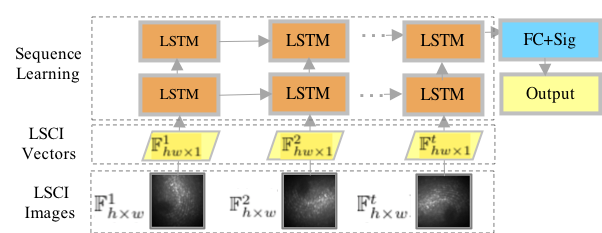}
    \caption{  LSCI-LSTM network fed by a sequence of the vectorized images of the LSCI-2D frames.  }
    \label{fig:LSTM_pipe}
\end{figure}

\section{Experimental Evaluation} 
\label{sec:NumericalResults}
FPAD classification performance is evaluated, per fold, for each of the studied architectures (BaseN, ResN, IncpN, Conv3, and LSTM) in terms of the metrics described in \secref{sec:Metrics}. The mean and standard deviation (std) of the metrics are computed over the total number of the folds for each of the partitioning strategies. Each of the studied architectures has been tuned separately for the spatial and temporal patch-sizes within the following ranges: $h\times w \in [8\times 8,16\times16,32\times 32,64\times64]$ for the  spatial-patch sizes and $t \in [5,10,50,100]$ for the temporal-patch-sizes. 

The averaged-ROC curves and means and stds for the three-fold partitioning are shown in \figref{fig:ROC3folds} and \tabref{tab:3fold_Results}, respectively. The performance of the LSTM architecture supersedes other studied architectures for FPAD, which is intuitive since the LSTM learn the inherent temporal dynamics of bona fide presentations and presentation attacks.

Similarly, ROC curves and the means and stds for the LOAO strategy are shown in Figures \ref{fig:ROCBaseN}-\ref{fig:ROCLSTM} and \tabref{tab:LOOResults}, respectively. Overall, it can be seen that the LSTM again outperforms other architectures. By inspecting the results of the different folds of the LOAO strategy at Figures \ref{fig:ROCBaseN}-\ref{fig:ROCLSTM}, it can be seen that the most challenging fold, which has the worse performance by each of the networks, is Fold\#5 corresponding to the dragon-skin attack.

\tabref{tab:TunePatchSizeLOO} shows the effect of varying patch-sizes for the LSTM network, which has the best FPAD performance. The results indicate that the performances are reduced by increasing the spatial-patch-size from 8$\times$8 to  16$\times$16. Our justification for this behavior is that increasing the spatial-patch-size leads to an increase to the number of the weights to be trained by the network, which  needs having access to a larger training-dataset. On the other hand, given a fixed spatial-patch size ($h \times w$), it can be seen that expanding the temporal patch-size ($t$) improves the performance, possibly because the LSTM has access to more temporal samples considering this fact that within the LSTM network the weights are shared across time.
   
\begin{table*}[thp]
\centering
\footnotesize
\begin{tabular}{r cc c c c c cc} \toprule
\multicolumn{2}{c}{}      &    APCER      &    BPCER       &   ACER      &    BPCER20       &  TPR02     &     TPR0     &       AUC      \\ \midrule 
\multirow{ 1}{*}{BaseN} 
 &     & 0.125$\pm$0.070   & 0.008$\pm$0.008   & 0.066$\pm$0.031  &  0.032$\pm$0.027  &  0.872$\pm$0.056  &  0.839$\pm$0.065   &  0.989$\pm$0.008   \\
\multirow{ 1}{*}{ResN}     
&     &\textbf{0.091$\pm$0.039}  &  0.033$\pm$0.038  &  0.062$\pm$0.027  &  0.067$\pm$0.060 &   0.823$\pm$0.089  &  0.754$\pm$0.092  &    0.989$\pm$0.010    \\
 
\multirow{ 1}{*}{IncpN}    
&     &0.098$\pm$0.080  &  0.103$\pm$0.124  &  0.100$\pm$0.050  &  0.057$\pm$0.046  &  0.808$\pm$0.060  &  0.739$\pm$0.104   &   0.985$\pm$0.012   \\
\multirow{ 1}{*}{Conv3}     
 &     & 0.134$\pm$0.055   &  \textbf{ 0.005$\pm$0.006}   &   0.069$\pm$0.026  &    0.023$\pm$0.008  &    0.839$\pm$0.057    &  0.816$\pm$0.072     &  0.991$\pm$0.006   \\  
\multirow{ 1}{*}{LSTM}      
&     & 0.097$\pm$0.042    &  0.006$\pm$0.005    &  \textbf{0.052$\pm$0.020 }   &  \textbf{0.019$\pm$0.015}   &  \textbf{ 0.881$\pm$0.041 }    &\textbf{ 0.858$\pm$0.049 }    &  \textbf{0.992$\pm$0.006 }   \\  \midrule
\end{tabular}
\caption{PAD classification results of the BaseN, ResN, IncpN, Conv3, and LSTM networks. Results are reported in terms of APCER, BPCER, ACER, BPCER20, TPR02 and AUC metrics at different columns. The reported values are in percentages and reflect the average and std  of the metric computed over the  3-folds of the 3-Fold partitioning (\secref{sec:DataPartitioning}). 
}  
\label{tab:3fold_Results}
\end{table*}

\begin{figure*}
    \centering
    \subfigure[Averaged ROCs of the 3-Fold]{\label{fig:ROC3folds}\includegraphics[width=4.2cm]{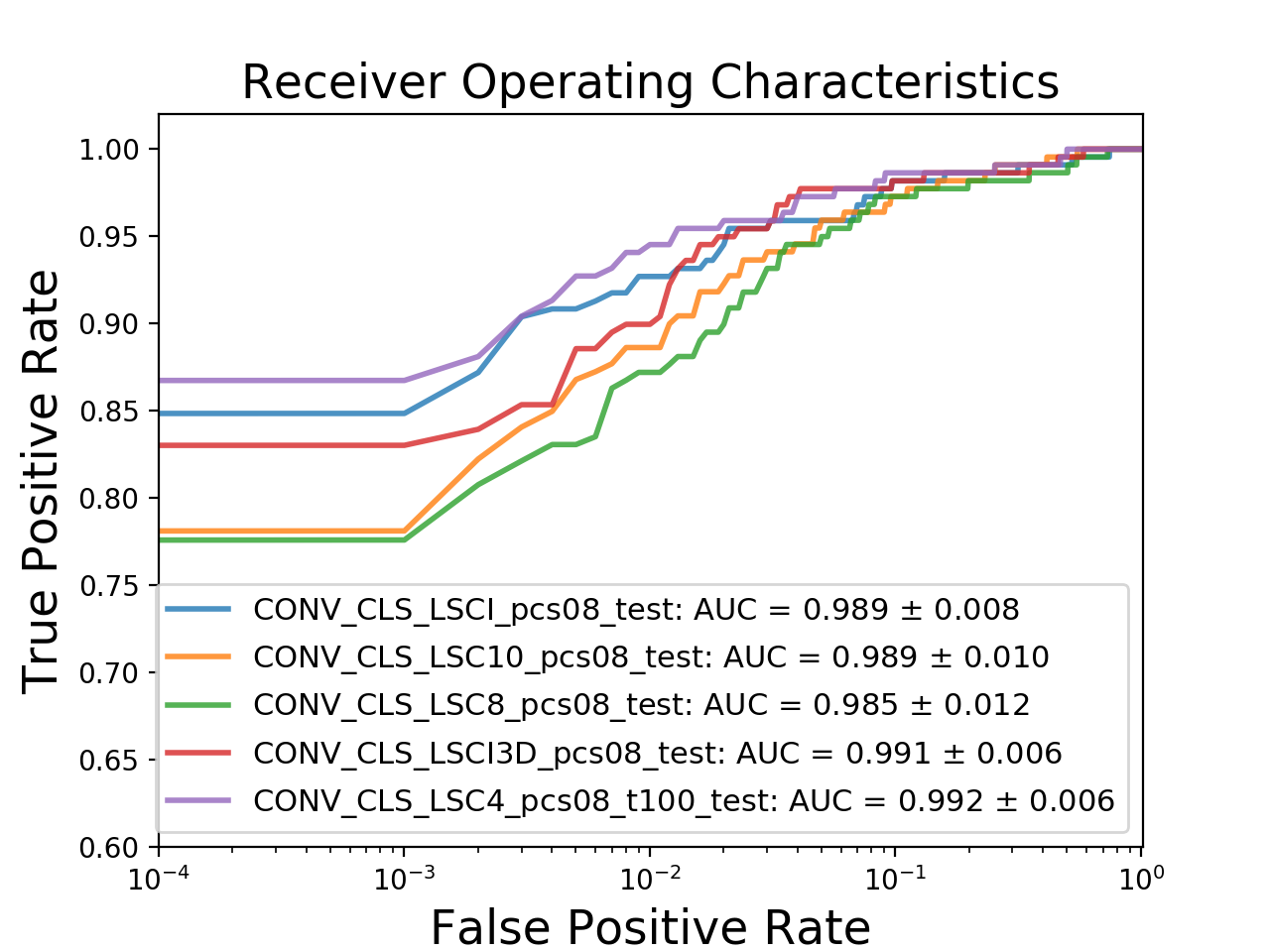}}
    \subfigure[ROCs of BaseN per fold of LOAO ]{\label{fig:ROCBaseN}\includegraphics[width=4.2cm]{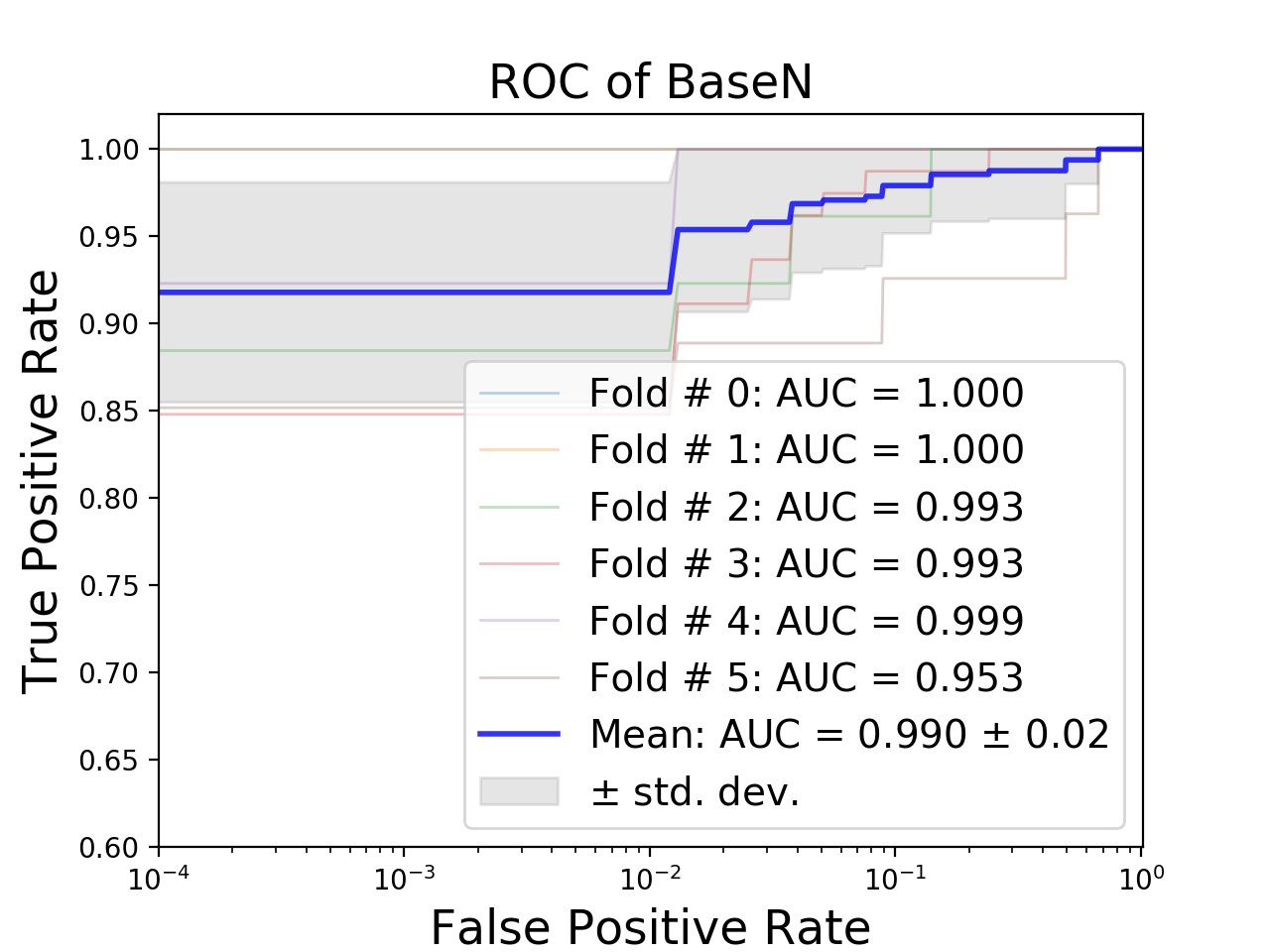}}
    \subfigure[ROCs of ResN per fold of LOAO]{\label{fig:ROCResN}\includegraphics[width=4.2cm]{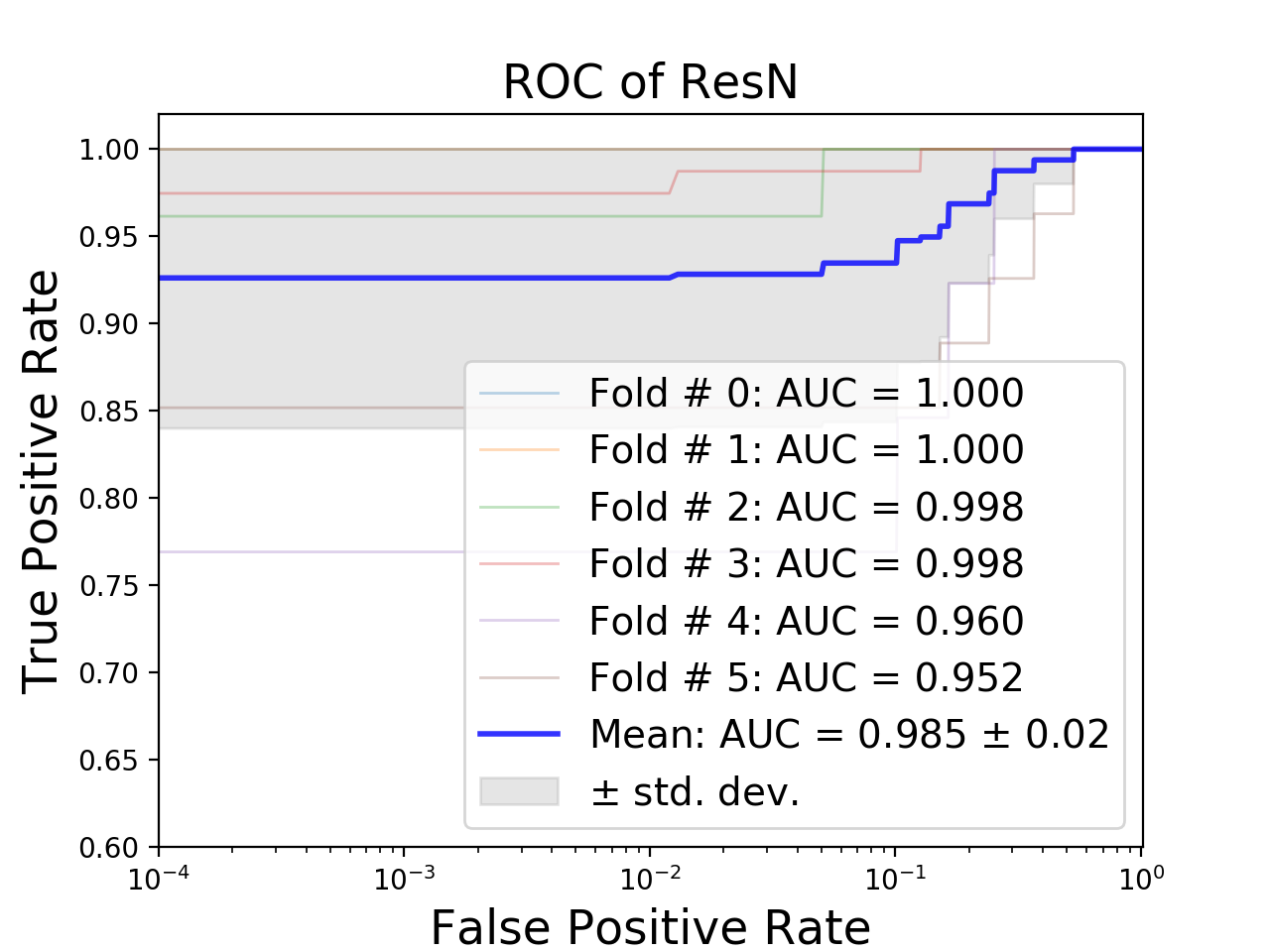}} \\ [-2.6ex]
    \subfigure[ROCs of IncpN per fold of LOAO]{\label{fig:ROCIncpN}\includegraphics[width=4.2cm]{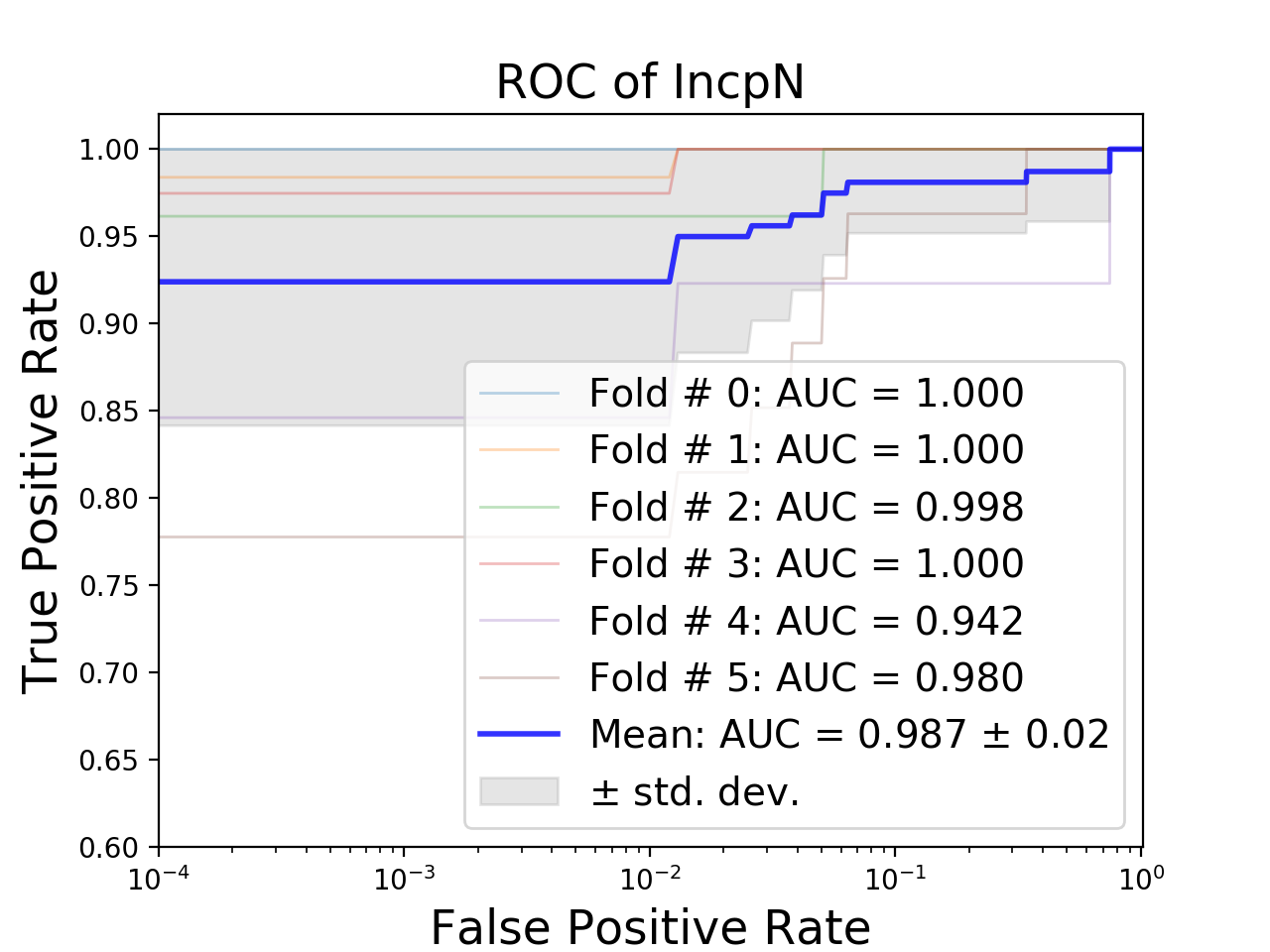}}
    \subfigure[ROCs of Conv3 per fold of LOAO]{\label{fig:ROCConv3}\includegraphics[width=4.2cm]{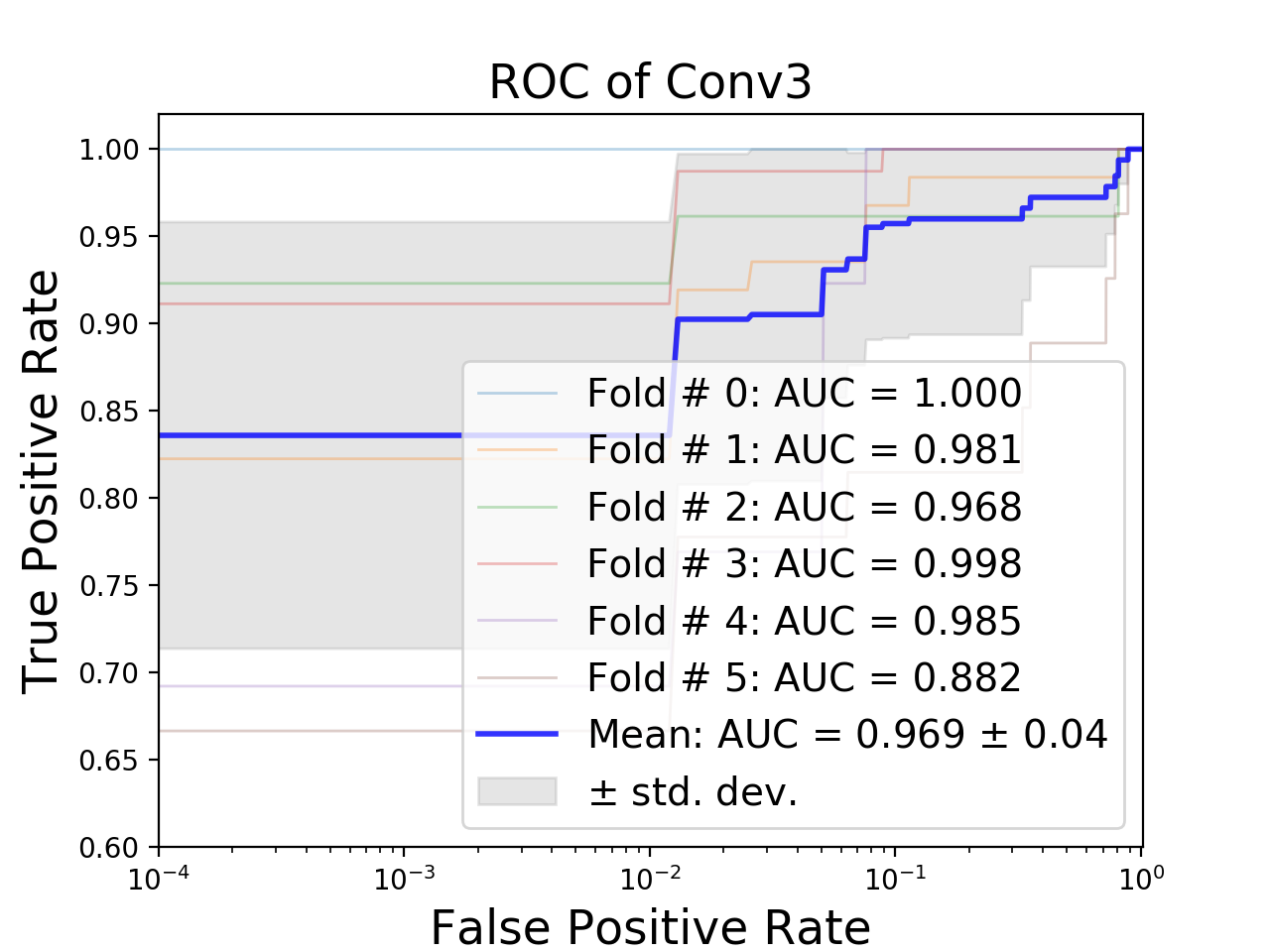}}
    \subfigure[ROCs of LSTM per fold of LOAO]{\label{fig:ROCLSTM}\includegraphics[width=4.2cm]{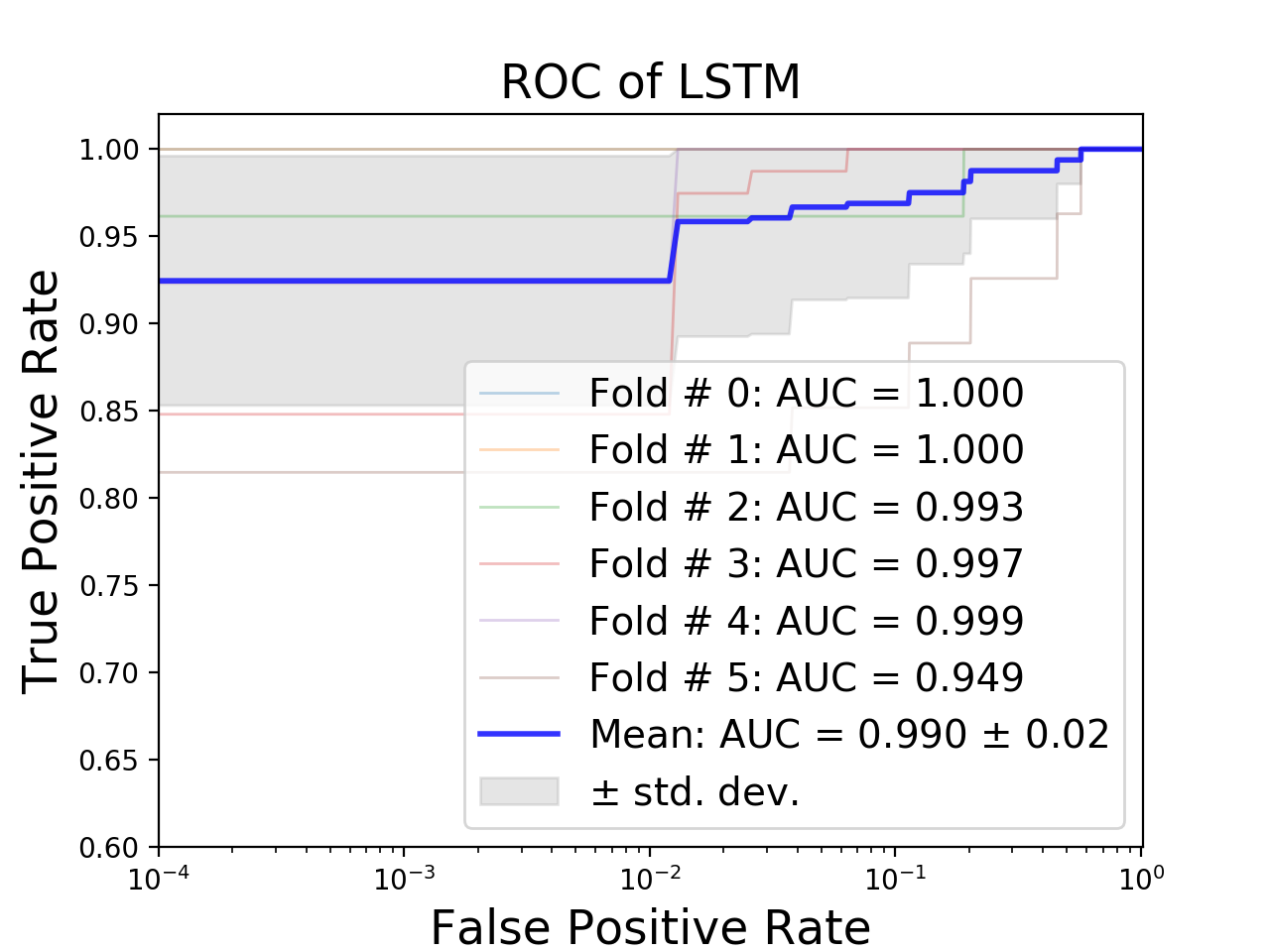}}
    \caption{ROC  curves of the different partitioning. (a) Averaged ROC curves of the networks based on the 3-Fold partitioning. (b-f) ROC curves of the networks per-fold of the LOAO partitioning and averaged ROCs. }
    \label{fig:ROCs}
\end{figure*}

\begin{table*}[thp]
\centering
\footnotesize
\begin{tabular}{r cc c c c c c} \toprule
\multicolumn{2}{c}{}      &    APCER      &    BPCER       &   ACER      &    BPCER20       &  TPR02        &       AUC      \\  \midrule
\multirow{ 1}{*}{BaseN} 
&    & 0.139$\pm$0.121&    0.002$\pm$0.005&    0.070$\pm$0.059&    0.097$\pm$0.178&    0.918$\pm$0.063&     \textbf{0.990$\pm$0.017}    \\   
\multirow{ 1}{*}{ResN}     
&    & 0.220$\pm$0.330&    \textbf{0.000$\pm$0.000}&    0.110$\pm$0.165&    0.103$\pm$0.150&    \textbf{0.926$\pm$0.086}&      0.985$\pm$0.020     \\    
\multirow{ 1}{*}{IncpN}    
&    & 0.226$\pm$0.240&    \textbf{0.000$\pm$0.000}&    0.113$\pm$0.120&    0.135$\pm$0.275&    0.924$\pm$0.082&   0.987$\pm$0.021    \\   
\multirow{ 1}{*}{Conv3}     
&    & 0.165$\pm$0.155&    0.013$\pm$0.007 &   0.089$\pm$0.074 &   0.160$\pm$0.281 &   0.836$\pm$0.122&     0.969$\pm$0.040 \\  
\multirow{ 1}{*}{LSTM}      
&    & \textbf{0.092$\pm$0.093}&    \textbf{0.000$\pm$0.000}&    \textbf{0.046$\pm$0.047}&    \textbf{0.080$\pm$0.168}&      0.925$\pm$0.071&     \textbf{0.990$\pm$0.018}      \\  \toprule  
\end{tabular}
\caption{PAD classification results of the BaseN, ResN, IncpN, Conv3, and LSTM networks. Results are reported in terms of APCER, BPCER, ACER, BPCER20, TPR02 and AUC metrics at different columns. The reported values are in percentages and reflect the average and std of the metrics computed over the  6-folds of the LOAO partitioning (\secref{sec:DataPartitioning}).  }
\label{tab:LOOResults}
\end{table*}

\begin{table*}[thp]
\centering
\footnotesize
\begin{tabular}{lc cc ccc} \toprule
$h\times w\times t$&        APCER    &       BPCER        &           ACER     &         BPCER20   &            TPR02    &                         AUC     \\      \midrule
4x4x5           & 0.239$\pm$0.176    & \textbf{0.000$\pm$0.000}    & 0.119$\pm$0.088    & 0.175$\pm$0.240    & 0.897$\pm$0.094        & 0.981$\pm$0.026     \\  
4x4x10          & 0.286$\pm$0.286    & \textbf{0.000$\pm$0.000}    & 0.143$\pm$0.143    & 0.116$\pm$0.159    & 0.918$\pm$0.076        & 0.985$\pm$0.022      \\  
4x4x50          & 0.194$\pm$0.152    & \textbf{0.000$\pm$0.000}    & 0.097$\pm$0.076    & 0.072$\pm$0.133    & \textbf{0.941$\pm$0.070 }       & 0.990$\pm$0.018     \\  
4x4x100         & 0.143$\pm$0.127    & \textbf{0.000$\pm$0.000}    & 0.071$\pm$0.063    & 0.084$\pm$0.120    & 0.935$\pm$0.069       & 0.990$\pm$0.015     \\  \midrule
8x8x5           & 0.194$\pm$0.160    & \textbf{0.000$\pm$0.000}    & 0.097$\pm$0.080    & 0.105$\pm$0.186    & 0.846$\pm$0.134        & 0.987$\pm$0.018     \\  
8x8x10          & 0.192$\pm$0.160    & \textbf{0.000$\pm$0.000}    & 0.096$\pm$0.080    & 0.091$\pm$0.175    & 0.860$\pm$0.130     & 0.988$\pm$0.018      \\  
8x8x50          & 0.160$\pm$0.117    & \textbf{0.000$\pm$0.000}    & 0.080$\pm$0.059    & 0.093$\pm$0.185    & 0.912$\pm$0.071     & 0.988$\pm$0.020     \\  
8x8x100         & \textbf{0.092$\pm$0.093}    & \textbf{0.000$\pm$0.000}    & \textbf{0.046$\pm$0.047}    & 0.080$\pm$0.168    & 0.925$\pm$0.071        & 0.990$\pm$0.018      \\  \midrule
16x16x5           & 0.212$\pm$0.303    & 0.008$\pm$0.006    & 0.110$\pm$0.149    & 0.080$\pm$0.135    & 0.837$\pm$0.160         & 0.989$\pm$0.017    \\  
16x16x10          & 0.214$\pm$0.299    & 0.006$\pm$0.006    & 0.110$\pm$0.147    & 0.072$\pm$0.133    & 0.876$\pm$0.144        & 0.989$\pm$0.018     \\  
16x16x50          & 0.162$\pm$0.195    & 0.011$\pm$0.009    & 0.087$\pm$0.094    & 0.137$\pm$0.279    & 0.836$\pm$0.167         & 0.985$\pm$0.027    \\  
16x16x100         & 0.169$\pm$0.193    & 0.006$\pm$0.006    & 0.088$\pm$0.095    & 0.167$\pm$0.334    & 0.878$\pm$0.104       & 0.982$\pm$0.035    \\  \midrule
32x32x5           & 0.235$\pm$0.171    & 0.002$\pm$0.005    & 0.119$\pm$0.084    & 0.078$\pm$0.109    & 0.842$\pm$0.145      & 0.989$\pm$0.013     \\  
32x32x10          & 0.191$\pm$0.156    & 0.002$\pm$0.005    & 0.097$\pm$0.077    & \textbf{0.044$\pm$0.067}    & 0.873$\pm$0.128        & 0.994$\pm$0.010    \\  
32x32x50          & 0.148$\pm$0.125    & 0.000$\pm$0.000    & 0.074$\pm$0.063    & 0.116$\pm$0.188    & 0.907$\pm$0.099      & 0.987$\pm$0.024    \\  
32x32x100         & 0.220$\pm$0.188    & 0.000$\pm$0.000    & 0.110$\pm$0.094    & 0.053$\pm$0.087    & 0.885$\pm$0.136    & \textbf{0.993$\pm$0.013}  \\    \toprule
\end{tabular}
\caption{Impact of varying patch-sizes $h\times w\times t$ (first column) on the  classification results of the LSTM network for the LOAO partitioning.   The values represent the mean and std of the metrics computed over  the  6-folds of the LOO partitioning (\secref{sec:DataPartitioning}).}  
\label{tab:TunePatchSizeLOO}
\end{table*}

\section{ Conclusions}
\label{sec:Conclusions}
Towards deeper analysis of the LSCI modality and its effectiveness, we collected a dataset consisting of 3961 LSCI images (3743 bona fide and 218 PA cases), including six different attack-types  (conductive paper, conductive silicone, transparency, silicone-I, silicone-II, and dragon-skin), collected from 335 unique subjects. We applied a variety of deep neural network architectures, including spatial 2D and 3D convolutions, and a combination of both using LSTM modules. All the networks were tuned for the spatial and temporal patch-sizes. We performed our validations following a 3-Fold partitioning strategy. Moreover, we deployed a leave-one-attack-out strategy due to the importance of assessing the ability to detect an unseen attacks. Validation results indicates that the best FPAD classification performance was achieved by LSTM network. Further inspection of the results shows that dragon-skin overlays constitute the most challenging attack in our dataset. Investigating the reasons behind the difficulty of dragon skin attacks for LSCI is certainly part of our future work. We are planning on collecting more data with much more PA sample counts and variety.
Furthermore, we consider the fusion among different models and other modalities towards reaching higher FPAD performance, particularly, for unseen attacks.

\section*{Acknowledgment}
This research is based upon work supported by the Office
of the Director of National Intelligence (ODNI), Intelligence
Advanced Research Projects Activity (IARPA),
via IARPA R\&D Contract No. 2017-17020200005. The
views and conclusions contained herein are those of the authors
and should not be interpreted as necessarily representing
the official policies or endorsements, either expressed
or implied, of the ODNI, IARPA, or the U.S. Government.
The U.S. Government is authorized to reproduce and distribute
reprints for Governmental purposes notwithstanding
any copyright annotation thereon.

{\small
\bibliographystyle{}
\bibliography{biblio}
}

\end{document}